\documentclass{article}

\usepackage{arxiv}

\usepackage[utf8]{inputenc} % allow utf-8 input
\usepackage[T1]{fontenc}    % use 8-bit T1 fonts
\usepackage{hyperref}       % hyperlinks
\usepackage{url}            % simple URL typesetting
\usepackage{booktabs}       % professional-quality tables
\usepackage{amsfonts}       % blackboard math symbols
\usepackage{nicefrac}       % compact symbols for 1/2, etc.
\usepackage{microtype}      % microtypography
\usepackage{lipsum}
\usepackage{graphicx}
\usepackage{multirow}

\title{Continual Learning with Columnar Spiking Neural Networks}

\author{
Denis Larionov \\
  Chuvash State University, \\
  Cheboksary, Russia\\
  Cifrum, Moscow, Russia\\
  \texttt{denis.larionov@gmail.com} \\
  \And
Nikolay Bazenkov \\
  Trapeznikov Institute of Control Sciences, \\
  Moscow Institute of Physics and Technology, \\
  Moscow, Russia \\
  \And
Mikhail Kiselev \\
  Chuvash State University, \\
  Cheboksary, Russia\\
}

\begin{document}

\maketitle

\begin{abstract}
Continual learning is a key feature of biological neural systems, but artificial neural networks often suffer from catastrophic forgetting. Instead of backpropagation, biologically plausible learning algorithms may enable stable continual learning. This study proposes columnar-organized spiking neural networks (SNNs) with local learning rules for continual learning and catastrophic forgetting. Using CoLaNET (Columnar Layered Network), we show that its microcolumns adapt most efficiently to new tasks when they lack shared structure with prior learning. We demonstrate how CoLaNET hyperparameters govern the trade-off between retaining old knowledge (stability) and acquiring new information (plasticity). We evaluate CoLaNET on two benchmarks: Permuted MNIST (ten sequential pixel-permuted tasks) and a two-task MNIST/EMNIST setup. Our model learns ten sequential tasks effectively, maintaining 92\% accuracy on each. It shows low forgetting, with only 4\% performance degradation on the first task after training on nine subsequent tasks. %We found that among the hyperparameters being investigated, the adaptive spiking threshold significantly influences the network performance.
\end{abstract}


\section{Introduction}
\label{sec:intro}

Living organisms adapt continuously to changing environments. They accumulate, update, and utilize knowledge throughout their lives. Artificial intelligence systems however still face difficulties acquiring new knowledge and skills on top of the existing ones. Continual learning (CL) \cite{Ring1997}, often called incremental or lifelong learning \cite{Kudithipudi2022}, addresses this challenge in artificial intelligence.

Modern artificial neural networks (ANNs) assume stationary data distributions throughout training. In CL, this assumption fails -- models must handle dynamically changing input distributions across sequential tasks. Training on new tasks often severely degrades performance on previous tasks. This phenomenon is known as catastrophic forgetting (CF) \cite{McCloskey1989, Ratcliff1990}. Backpropagation causes CF by requiring global parameter updates when learning new tasks.

Various approaches address CF in deep learning (Section \ref{sec:related_works}). Biologically inspired methods use local learning, temporal dynamics, neuronal competition, and modulated plasticity. Spiking neural networks (SNNs) \cite{Kiselev2020a} provide a mathematical implementation of these principles.

However, \cite{Antonov2022} shows that local learning alone in SNNs is insufficient. The CF solution may require higher abstraction levels, considering functionally organized neuron groups rather than individual neurons. Cortical microcolumns exemplify such organization. CoLaNET architecture \cite{Kiselev2024a, Kiselev2024b, Kiselev2025b} presents an SNN based on columnar organization principles, combining columnar structure with local learning and other distinctive features. This combination should theoretically enable CL capabilities.

We investigate MNIST image classification using two experimental sets. First, we use the Permuted MNIST protocol from~\cite{Goodfellow2013} to generate independent CL tasks. Second, we use a two-task protocol with Extended MNIST (EMNIST) \cite{Cohen2017}, adding handwritten letters to classical MNIST. We compare against a 1-layer ANN baseline with 512 hidden neurons (similar to \cite{Goodfellow2013}).

In the Permuted MNIST ten-task experiment, CoLaNET shows perfect memory stability but sacrifices plasticity. By tuning microcolumn numbers, silent synapses, and adaptive threshold, CoLaNET achieves both low forgetting (4.35\% across 10 tasks) and decent average accuracy (92.31\%). In the E/MNIST experiment, CoLaNET does not perform well due to shared features between tasks. However, when applied to datasets with random permutations (like Permuted MNIST), CoLaNET outperforms other approaches. The source code and experiment results are open-source and available at: https://gitflic.ru/project/dlarionov/cl.

\section{Related works}
\label{sec:related_works}

Several surveys cover continual learning research \cite{Ven2024, Wang2024}. Most recent studies focus on computer vision classification tasks \cite{DeLange2022}. Fewer works address text-based tasks \cite{Biesialska2020}, large language models \cite{Wu2024}, and reinforcement learning \cite{Khetarpal2022}. Common CL approaches fall into several classes.

\emph{Replay}. Replay strategies approximate past task data distributions by storing training samples in memory buffers \cite{Chaudhry2019} or training generative models to reproduce past data. Interleaving old and new data during training helps retain knowledge of earlier tasks while learning new ones.

\emph{Regularization}. These methods add penalty terms to preserve important parameters or behaviors from past tasks. Weight regularization restricts changes to critical parameters \cite{Kirkpatrick2016, Zenke2017}, while function regularization maintains stable input-output mappings.

\emph{Optimization manipulation}. This strategy adapts optimization itself for CL. Examples include seeking flatter minima through optimizer choice or learning rate scheduling \cite{Mirzadeh2020}, using adaptive per-synapse learning rates, projecting gradients to avoid interference \cite{Chaudhry2018b}, or employing fast/slow weights to separate short- and long-term updates.

\emph{Architectural modifications.} These approaches dynamically adjust network structure for new tasks. Multi-head architectures assign separate output heads per task, while masking protects task-specific weights. Alternatives include gating mechanisms or dynamic expansion to isolate or partition knowledge.

\emph{Template-based classification.} This approach uses class prototypes or generative models, creating templates (e.g., embedding-space prototypes) for each class and classifying new samples by proximity \cite{DeLange2022}.

CF extends beyond mere forgetting \cite{Kudithipudi2022}. First, CF represents the plasticity-stability dilemma: excessive learning plasticity interferes with memory stability, and vice versa \cite{Grossberg1982}. Second, it relates to generalizability—distinguishing data distribution differences within and across tasks. Third, CF can be viewed as a computational resource optimization problem when training on new tasks. While reusing all past task data for new tasks would eliminate CF, it introduces high computational and storage costs along with potential privacy concerns. Training inefficiencies may also arise from excessive repetitions on single tasks (epochs). Thus, CL approaches typically restrict training to one epoch per task \cite{Yin2021}.

Nearly all successful CL strategies rely on gradient-based optimization, which causes CF. Local learning based on Hebbian plasticity offers an alternative. Enhanced with temporal dynamics, competition, gating mechanisms, and modulated plasticity, it provides a rich CL framework. Spiking neural networks (SNNs) \cite{Kiselev2020a} commonly implement these principles.

Relatively few studies explore SNNs for CL. Section \ref{sec:comparison} provides detailed comparison with \cite{Antonov2022}.
\section{Continual learning framework}

\subsection{Problem definition}
\label{sec:problem}

Continual learning handles dynamically changing data distributions and task conditions. A model with parameters $\theta$ learns sequential tasks while preserving performance on older tasks after learning new ones \cite{Wang2024}.

Formally, training data for problem $t$ is $\mathcal{D}_{t,b} = \{\mathbf{X}_{t,b}, \mathbf{Y}_{t,b}\}$ where $\mathbf{X}_{t,b}$ is input data, $\mathbf{Y}_{t,b}$ are labels, $t \in \mathcal{T} = \{1, \dots, k\}$ is the task index, and $b \in \mathcal{B}_t$ is the batch index. Here $\mathcal{T}$ is the task space and $\mathcal{B}_t$ is the batch space for task $t$. Tasks are defined by their training data $\mathcal{D}_t$. Data distribution within task $\mathcal{D}_t := p(\mathbf{X}_t, \mathbf{Y}_t)$ remains consistent between train and test sets. Labels $\mathbf{Y}_t$ and task index $t$ may be unavailable during training.

\subsection{CL scenarios}
\label{sec:scenarios}

The taxonomy in \cite{Ven2024} categorizes CL scenarios using these concepts:

\emph{Task-based and task-free.} Task-based problems present discrete tasks sequentially with clear boundaries. Task-free scenarios have discrete tasks but soft transitions.

\emph{Task-, domain-, and class-incremental.} According to \cite{Ven2019} task-incremental learning (TIL) learns distinct tasks with always-available task identifiers. Domain-incremental learning (DIL) maintains a consistent task structure (label space) but changes input data context over time, making context determination unnecessary. Class-incremental learning (CIL) has disjoint label spaces per task, with task identifiers available only during training.

\emph{Streaming.} Each example appears only once.

\emph{Online.} Only one example appears at a time.

The experiments conducted in the study belong to the class of online streaming task-based domain incremental learning.

\subsection{Evaluation metrics}
\label{sec:metrics}

\emph{Overall performance.} We evaluate overall performance using average accuracy (AA) and average incremental accuracy (AIA). Let \(a_{k,j} \in [0, 1]\) be classification accuracy evaluated on the test set of task j after incremental learning of the k-th task (j $\leq$ k). Average accuracy is:

\begin{equation}
AA_k = \frac{1}{k} \sum_{j=1}^k a_{k,j}.
\label{eq:aa}
\end{equation} 

Average incremental accuracy is:

\begin{equation}
AIA_k = \frac{1}{k} \sum_{i=1}^k AA_i.
\label{eq:aia}
\end{equation} 

AA reflects current overall performance, while AIA aggregates performance across previous tasks.

\emph{Memory stability.} We evaluate memory stability using forgetting measure (FM) and backward transfer (BWT)~\cite{Chaudhry2019}. FM of task $j$ after training on task $k$ is the difference between maximum past performance and current performance:

\begin{equation}
f_{j,k} = \max_{i \in \{1, \dots, k-1\}} (a_{i,j} - a_{k,j}), \quad \forall j < k.
\label{eq:stability}
\end{equation} 

FM of task $k$ is average forgetting across previous tasks: 

\begin{equation}
FM_k = \frac{1}{k-1} \sum_{j=1}^{k-1} f_{j,k}.
\label{eq:fm}
\end{equation} 

BWT is the average influence of learning task $k$ on previous tasks:

\begin{equation}
BWT_k = \frac{1}{k-1} \sum_{j=1}^{k-1} (a_{k,j} - a_{j,j}),
\label{eq:btw}
\end{equation}

\noindent where negative BWT values indicate forgetting.

\emph{Generalization ability.} We evaluate generalization using forward transfer (FWT), the average influence of previous tasks on task $k$:

\begin{equation}
FWT_k = \frac{1}{k-1} \sum_{j=2}^k (a_{j,j} - \tilde{a}_j),
\label{eq:fwt}
\end{equation}

\noindent where \(\tilde{a}_j\) is baseline model accuracy on task j.

%The parameter \(a_{k,j}\), in addition to classification, can also be adapted to other types of tasks, such as average precision (AP) for object detection \cite{Shmelkov2017}, Intersection-over-Union (IoU) for semantic segmentation \cite{Michieli2019}, Frechet Inception Distance (FID) for image generation \cite{Wu2018}, and normalized reward for RL \cite{Ahn2019}.

\section{Columnar spiking network}
\label{sec:colanet}

CoLaNET (Columnar Layered Network) architecture \cite{Kiselev2024a, Kiselev2024b, Kiselev2025b} embodies columnar organization in SNNs with local learning, neuronal competition, modulated plasticity, and gating mechanisms. Designed for supervised classification, CoLaNET's key feature integrates anti-Hebbian plasticity (degrading weights) with modulated plasticity (strengthening weights), counteracting degradation and enabling effective learning.

\subsection{Network architecture}
\label{sec:architecture}

CoLaNET consists of multiple identical modules (columns), with column count matching class count. Each column has five layers: trainable input, three intermediate processing, and output layer. The lower three layers organize into microcolumns - small interconnected neuron groups recognizing subclasses within single classes. All neurons are described by the simplest LIF (leaky integrate-and-fire) model with slight modifications described later.

\begin{figure}[htbp]
  \centering  
  \includegraphics[width=0.9\textwidth]{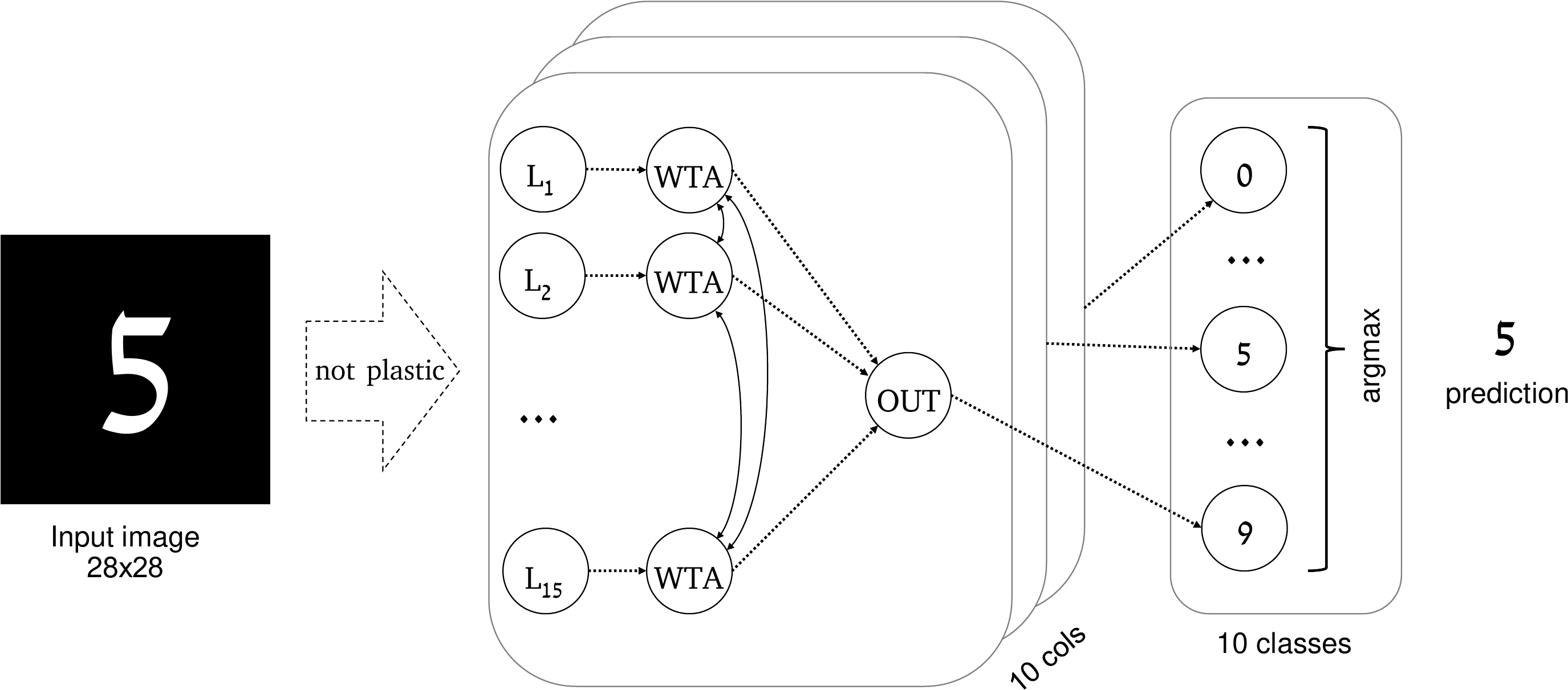}
  \caption{CoLaNET architecture, inference regime. The image is presented over 10 time steps, followed by 10 time steps of silence (empty image). The neuron L that first generates a spike activates the WTA neuron, which then suppresses other WTA neurons within the same column and propagates the spike to the OUT neuron. The class is determined by the column producing the highest number of spikes.}
  \label{fig:inference}
\end{figure}

CoLaNET converts pixel intensities to spike sequences using rate coding, where spike generation probability at each time step equals pixel intensity ratio (0 intensity = zero probability, 255 = maximum probability). Images appear for 10 time steps, followed by 10 silence steps.

In the inference regime, assuming that all neurons have correct values of synaptic weights, CoLaNET operates as follows Figure \ref{fig:inference}. If the L neurons have the right values of their synaptic weights, then only an L neuron belonging to the correct column fires during image presentation. It causes firing of the WTA neuron in its microcolumn that in its turn forces the correct OUT neuron to fire.

\begin{figure}[htbp]
  \centering  
  \includegraphics[width=0.9\textwidth]{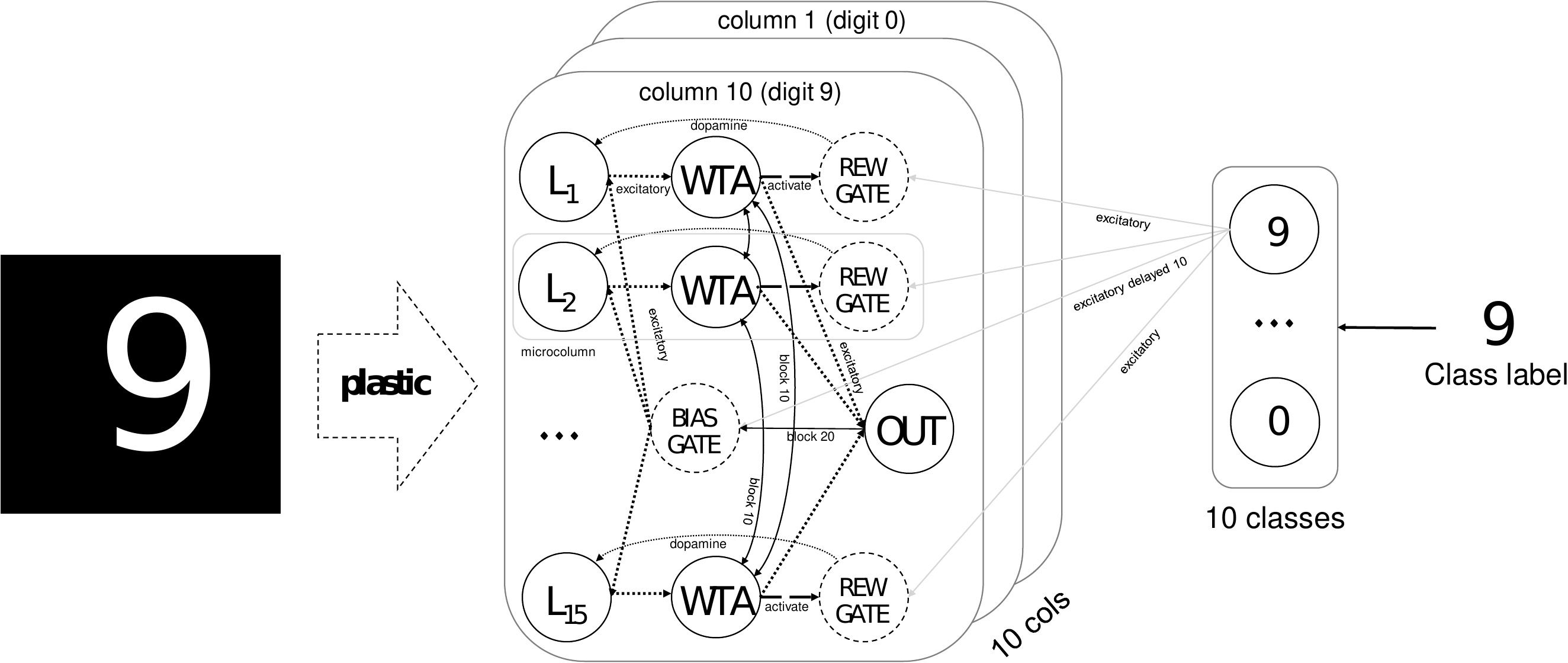}
  \caption{CoLaNET architecture, learning regime. Learning is governed by three factors: anti-Hebbian plasticity (weakening synaptic weights), dopamine-modulated plasticity (strengthening weights), and periodic synaptic renormalization. The label is presented for 20 time steps. Each class stimulates its own column.}
  \label{fig:training}
\end{figure}

CoLaNET training (Figure \ref{fig:training}) relies on microcolumn competition for activation when presented with input patterns. In the learning regime, the network obtains spikes encoding the current image as well as the information about class (label) of this image. The latter has the form of permanent activity of the respective image label input node (Figure \ref{fig:training} right). 

At the beginning of the learning process, all weights of all plastic synapses (of L neurons) are zero. Therefore, the stimulation from the input cannot make them fire. However, the network has another source of spikes – one of the class label nodes. It sends spikes to the REWGATE neurons of its column (see Figure \ref{fig:training}). However, these neurons are in the inactive state (see the discussion below about the active/inactive neuron states) and are not able to fire. Besides that, this node sends spikes to strong excitatory synapse of the BIASGATE neuron in its column. The connection between it and this neuron is slow – the spikes pass it for 10 time steps. Therefore, the first spike from the train emitted by the class label node reaches BIASGATE when the input stimulation has ended. This spike train induces constant firing of BIASGATE. The BIASGATE neuron is connected with the learning neurons by the excitatory synapse with the weight sufficient to force neuron to fire just before the end of the silence period (by the constant stimulation from BIASGATE). Since at the beginning, all learning neurons are identical, all of them fire simultaneously (only in the stimulated column, of course). All of them send powerful stimulation to the WTA neurons. But the WTA layer in one column (WTA means “winner takes all”) is designed so that no more than one neuron can fire simultaneously. Some random WTA neuron fires. It activates the REWGATE neuron in its microcolumn. But the REWGATE neurons still obtain stimulation from the class label node. Therefore, one REWGATE neuron immediately fires. It emits so-called dopamine spike coming at the special dopamine synapse of the learning neuron in the same microcolumn. This spike triggers the dopamine plasticity process. The dopamine plasticity rule says that all plastic synapses having obtained spikes some time before neuron firing are potentiated if the neuron receives a dopamine spike shortly after that firing. As a result of this process, one learning neuron in the column corresponding to the class presented gets slightly potentiated synapses connected to the recently active input nodes. These weights are still insufficient for firing solely from the input stimulation. However, the next time when a similar stimulus is presented, this winning neuron will have a positive value of the membrane potential at the beginning of stimulation from the BIASGATE. Therefore, it will have high chances to become a winner again, thus further potentiating the same set of synapses.

After some number of the plasticity acts described above, some L neurons acquire the ability to fire in response to input stimulation without help of BIASGATE neurons. In this case, the WTA neuron connected to the firing L neuron stimulates the OUT neuron of this column. It fires and blocks the BIASGATE neuron for the whole period of current object presentation (including the silence period) because stimulation from BIASGATE is not needed now.

This scheme has protection against wrong L neuron firing. It is implemented as a combination of two synaptic plasticity components – anti-Hebbian plasticity and dopamine plasticity. Dopamine plasticity was briefly described above. The anti-Hebbian plasticity mechanism is also simple. Whenever the neuron fires, all its plastic synapses having received a spike shortly before this are depressed. The correct stimuli are marked by the activity of the respective class label node which causes dopamine reward of the L neuron. If an L neuron fired and did not receive the dopamine reward, it fired wrongly and, therefore, the synapses which forced it to fire should be suppressed.

It remains to say about active and inactive neuron states, the only non-standard feature in the LIF model described (in the neurons WTA, REWGATE and BIASGATE). In the active state, a neuron behaves like a normal LIF neuron. In the inactive regime, presynaptic spikes do not change value of the membrane potential. The current neuron regime is determined by the sign of the neuron state component called the activity time a. The neuron is active if a > 0. This value can be modified (set to a negative or positive value) by spikes coming to special gating synapses. In their absence, it linearly changes toward 0. This feature is convenient for implementation of the neuron blocking/activation logics described above.

In [9], genetic algorithm hyperparameter optimization yields a CoLaNET configuration with 15 separate networks for MNIST classification. Each network contains 10 columns (matching class count) with 15 microcolumns each. Total neuron count is 7,050, with 2,250 having plastic synapses. This configuration achieves 95\% accuracy in single-epoch training, while individual ensemble networks reach only 75\%.

\subsection{CL features}
\label{sec:features}

CoLaNET’s adaptive firing threshold balances learning plasticity and memory stability \cite{Kiselev2024a}. The adaptive threshold $u_{tr}$ combines a constant component with a term proportional to positive weight sum: 

\begin{equation}
u_{tr} = u_{const} +  \alpha \sum_{i} w_i^+,
\label{eq:at}
\end{equation}

\noindent where $\alpha$ is the proportionality coefficient, $u_{const}$ is the fixed threshold part, and $\sum_{i} w_i^+$ is the positive weight sum ($w_i > 0$).

Parameter $\alpha$ controls neuron specialization degree for specific subclasses. Strong synapses increase firing threshold, enabling activation only for specific input patterns.

Synaptic resource renormalization \cite{Kiselev2024b} also influences neuron specialization. Strengthened synapses proportionally weaken others, and vice versa. CoLaNET disables renormalization by default but enables it by specifying virtual synapse count ($ns$) that absorbs parameter changes. When $ns=0$, renormalization effect maximizes; increasing $ns$ diminishes it to zero.

Microcolumn count per column also controls network capacity, crucial for scaling task numbers.

\subsection{ArNI-X framework}
\label{sec:arnix}

We implemented all SNN training and evaluation experiments using ArNI-X framework \cite{ArNI-X}. Since CoLaNET was originally designed in ArNI-X, the framework includes all necessary CoLaNET mechanics.

\section{Experiments}
\label{sec:experiments}

\subsection{Permuted MNIST}
\label{sec:experiments_pmnist}

\begin{figure}[htbp]
  \centering  
  \includegraphics[width=0.6\textwidth]{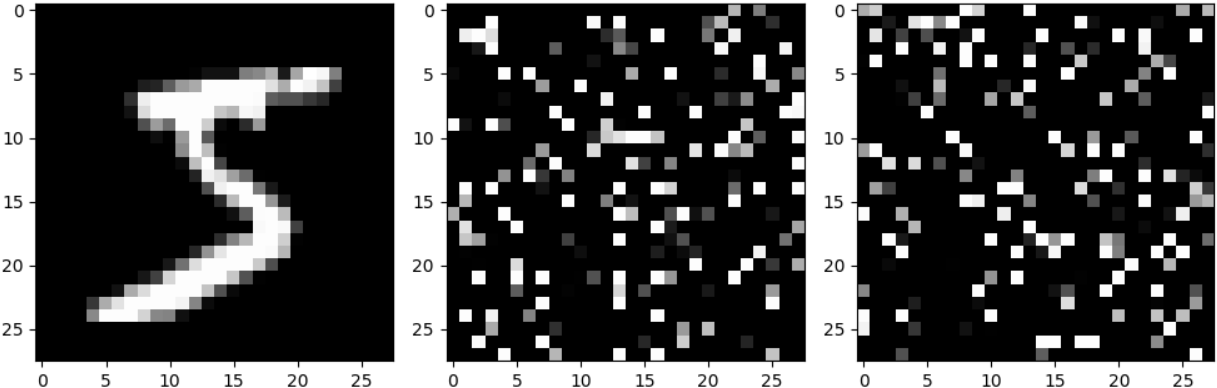}
  \caption{Two random pixel permutations of digit 5.}
  \label{fig:permutation}
\end{figure}

\emph{Tasks protocol.} Permuted MNIST is a popular CL framework with unlimited tasks \cite{Zenke2017, LopezPaz2017, Chaudhry2019}. Each task uses random pixel permutation of MNIST dataset. The permutation remains consistent across all task images. We evaluate models on all previous tasks~\cite{Goodfellow2013}. Figure \ref{fig:permutation} shows an example of permuted images of digit 5.

Random pixel permutation destroys spatial patterns, preventing CNN use and human classification. However, fully connected networks (like CoLaNET) find transformed images equally informative.

Sequential learning across ten MNIST permutations creates a degradation profile a 10×10 matrix where element \(a_{k,j} \in [0, 1]\) represents classification accuracy on task j's test set after learning task k (\(j \leq k\)). From this profile we compute Average Accuracy (AA), Average Incremental Accuracy (AIA), Forgetting Measure (FM) and Backward Transfer (BWT) (Section \ref{sec:metrics}).

\emph{Baseline model.} We trained a fully connected ANN with 512 neurons in the hidden layer and 10 neurons in the output layer for classification as a baseline model \cite{Larionov2024b}. The network uses SGD with adaptive learning rate (AdaDelta) and achieves 98\% accuracy on the MNIST test set after 5 epochs. For CL, we set one epoch per task. This slightly reduces individual task performance (98\% to 96\%) but reduces inter-task interference and maintains online learning.

Table \ref{tab:pmnist_ann} shows the baseline degradation profile. Columns represent tasks, rows represent training iterations on each task. Each cell contains test set classification accuracy for the respective task. Table \ref{tab:performance} presents final computed metrics.

\begin{table}[ht]
\centering
\caption{Baseline model performance. Columns are task identifiers, rows are training iterations.}
\label{tab:pmnist_ann}
\begin{tabular}{|c|*{10}{c|}}
\hline
\multirow{2}{*}{Iterations} & \multicolumn{10}{c|}{Tasks} \\
\cline{2-11}
 & 1 & 2 & 3 & 4 & 5 & 6 & 7 & 8 & 9 & 10 \\
 \hline
1 & \textbf{96.81} & 15.33 &  5.77 &  8.62 &  8.06 & 11.72 &  5.91 &  9.00 &  9.46 &  6.08 \\
\hline
2 & 94.82 & \textbf{96.52} &  4.77 &  6.68 & 14.72 & 12.16 &  6.35 &  9.91 &  9.47 &  6.92 \\
\hline
3 & 89.14 & 95.45 & \textbf{96.83} &  9.02 & 19.31 & 13.87 &  5.54 &  9.36 & 10.27 &  4.13 \\
\hline
4 & 81.73 & 93.93 & 95.58 & \textbf{96.50} & 16.28 & 14.07 &  7.37 &  9.98 &  9.50 &  6.96 \\
\hline
5 & 79.35 & 91.68 & 94.18 & 95.02 & \textbf{96.25} & 15.94 &  8.90 & 10.78 & 11.07 &  8.68 \\
\hline
6 & 71.54 & 86.79 & 92.13 & 91.67 & 95.10 & \textbf{96.18} &  7.99 &  9.12 &  9.67 &  8.90 \\
\hline
7 & 71.79 & 79.97 & 85.93 & 90.04 & 93.06 & 94.92 & \textbf{96.71} &  8.90 & 13.11 &  9.18 \\
\hline
8 & 67.39 & 73.21 & 80.01 & 84.26 & 89.49 & 92.07 & 95.16 & \textbf{96.02} & 13.64 &  9.73 \\
\hline
9 & 58.18 & 65.47 & 71.49 & 79.13 & 83.41 & 87.16 & 93.53 & 95.37 & \textbf{95.85} & 11.30 \\
\hline
10 & 49.43 & 55.05 & 60.97 & 70.72 & 76.77 & 80.90 & 87.52 & 93.46 & 93.93 & \textbf{95.45} \\
\hline
\end{tabular}
\end{table}

\emph{CoLaNET.} We implemented Permuted MNIST experiments with CoLaNET in three stages. Stage one uses the CoLaNET configuration from \cite{Kiselev2024b} (Section \ref{sec:colanet}), optimized for single-task performance without considering CL specifics like stability-plasticity balance.

We run ArNI-X sequentially across 10 Permuted MNIST tasks, then evaluate accuracy on all previous tasks (55 total runs per experiment). Each training iteration generates a network state file for evaluating all previous tasks. From iteration two onward, the network loads state from the previous task. CoLaNET experiments on dual Quadro RTX 6000 systems take 1h 20m for 15 microcolumns and 3h for 45 microcolumns.

Table~\ref{tab:colanet_dg1} shows results using optimal CoLaNET configuration from \cite{Kiselev2024b}: first-task accuracy meets expectations (94\%). Then significant performance drops occur: 10\% decrease by Task 2 (83\%), and 50\% by Task 3. Notably, the network shows zero forgetting: task accuracies remain constant throughout training.

\begin{table}[ht]
\centering
\caption{CoLaNET degradation profile for 15 microcolumns and $\alpha=0.023817$.}
\label{tab:colanet_dg1}
\begin{tabular}{|c|*{10}{c|}}
\hline
\multirow{2}{*}{Iterations} & \multicolumn{10}{c|}{Tasks} \\
\cline{2-11}
 & 1 & 2 & 3 & 4 & 5 & 6 & 7 & 8 & 9 & 10 \\
\hline
1 & \textbf{94.36} & & & & & & & & & \\
\hline
2 & 94.34 & \textbf{83.67} & & & & & & & & \\
\hline
3 & 94.34 & 83.61 & \textbf{34.17} & & & & & & & \\
\hline
4 & 94.34 & 83.61 & 34.68 & \textbf{36.75} & & & & & & \\
\hline
5 & 94.33 & 83.60 & 34.68 & 36.76 & \textbf{25.59} & & & & & \\
\hline
6 & 94.45 & 83.60 & 34.68 & 36.76 & 25.59 & \textbf{23.97} & & & & \\
\hline
7 & 94.45 & 83.60 & 34.68 & 36.76 & 25.59 & 23.97 & \textbf{12.01} & & & \\
\hline
8 & 94.43 & 83.60 & 34.68 & 36.76 & 25.59 & 23.97 & 12.03 & \textbf{15.25} & & \\
\hline
9 & 94.43 & 83.60 & 34.68 & 36.76 & 25.59 & 23.97 & 12.03 & 15.27 & \textbf{11.87}  & \\
\hline
10 & 94.43 & 83.60 & 34.68 & 36.76 & 25.59 & 23.97 & 12.03 & 15.27 & 11.87 & \textbf{13.55} \\
\hline
\end{tabular}
\end{table}

\begin{table}[ht]
\centering
\caption{Overall performance on Permuted MNIST}
\label{tab:performance}
\begin{tabular}{|c|c|c|c|c|}
\hline
 \multirow{2}{*}{Model} & \multicolumn{4}{c|}{Performance metrics} \\
 \cline{2-5}
& AA & AIA & FM & BWT \\
\hline
\textbf{Baseline model (1-layer ANN)} & & & &\\
avg & 88.80 & 29.39 & 9.17 & -9.17 \\
max & 96.81 & 47.11 & 22.10 & -22.10 \\
std & 6.19 & 12.05 & 6.53 & 6.53 \\
\hline
%\textbf{SNN CoLaNET} & AA & AIA & FM & BWT \\
%\hline
\textbf{CoLaNET (15 microcolumns, $\alpha=0.023817$)} & & & & \\
avg & 57.92 & 19.86 & 0.00 & 0.05 \\
max & 94.36 & 26.35 & 0.04 & -0.04 \\
std & 19.89 & 5.20 & 0.04 & 0.05 \\
\hline
\textbf{CoLaNET (15 microcolumns, $\alpha=0.005$)} & & & & \\
avg & 89.50 & 29.12 & 2.88 & -2.88 \\
max & 91.03 & 49.03 & 3.47 & -3.47 \\
std & 0.73 & 12.73 & 1.13 & 1.13 \\
\hline
\textbf{CoLaNET (45 microcolumns, $\alpha=0.1$)} & & & & \\
avg & 92.31 & 30.02 & 1.19 & -1.18 \\
max & 93.30 & 50.67 & 1.49 & -1.49 \\
std & 0.39 & 13.12 & 0.41 & 0.40 \\
\hline
\end{tabular}
\end{table}

Results show balance shifted toward memory stability at plasticity’s expense. The absence of forgetting suggests that microcolumns specialized for one task stop adapting to subsequent tasks entirely. After training on the first task, the network still retains a portion of microcolumns with low-magnitude weights capable of learning. However, this pool shrinks with each new task. Eventually the network cannot maintain high accuracy on novel tasks, with performance dropping to near random (10\%).

One mechanism for controlling the degree of neuron specialization is the adaptive threshold. It increases the activation threshold proportionally to the sum of positive weights, scaled by coefficient $\alpha$ (Equation~\ref{eq:at}). In CoLaNET's optimal configuration, $\alpha = 0.023817$ \cite{Kiselev2024b}.

\begin{figure}[htbp]
  \centering  \includegraphics[width=1\textwidth]{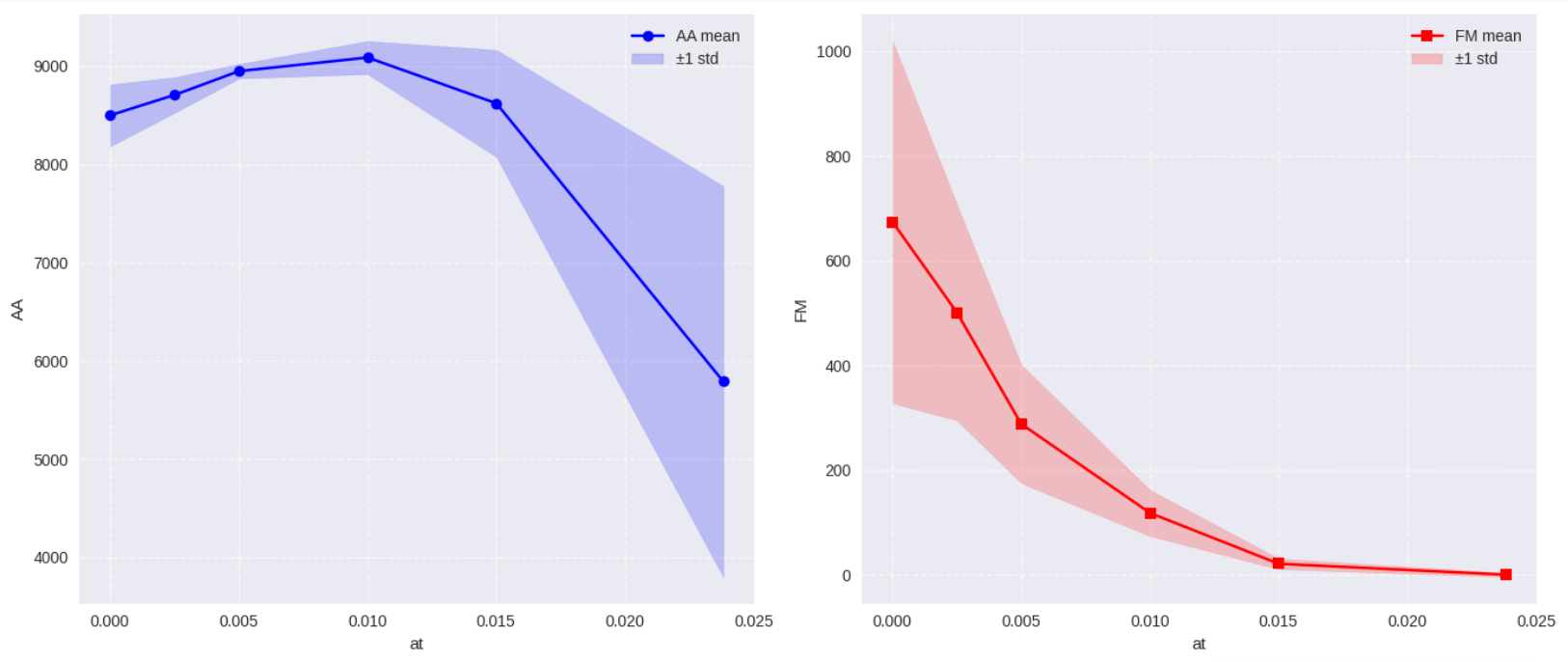}
  \caption{Average Accuracy and Forgetting Measure with different $\alpha$ for CoLaNET with 15 microcolumns per column.}
  \label{fig:at_metrics}
\end{figure}

Stage two uses varying $\alpha$ values from $0$ to $0.023817$. When $\alpha = 0$, the threshold loses adaptive properties, degrading classification accuracy from 94\% to 90\%. However, this enables full-capacity learning on each new task. Results again show skewed balance, now favoring learning plasticity over memory stability. Figure~\ref{fig:at_metrics} presents AA and FM metrics for different $\alpha$ values. For $\alpha > 0.015$, the network almost stops forgetting. Optimal value is $\alpha=0.005$, with degradation profile in Table~\ref{tab:colanet_dg2} and metrics in Table~\ref{tab:performance}. Learning plasticity degrades for $\alpha > 0.005$ as network capacity proves insufficient for new tasks. More microcolumns in CoLaNET configuration can improve this.

\begin{table}[ht]
\centering
\caption{CoLaNET degradation profile for 15 microcolumns and $\alpha=0.005$.}
\label{tab:colanet_dg2}
\begin{tabular}{|c|*{10}{c|}}
\hline
\multirow{2}{*}{Iterations} & \multicolumn{10}{c}{Tasks} \\
\cline{2-11}
 & 1 & 2 & 3 & 4 & 5 & 6 & 7 & 8 & 9 & 10 \\ \hline
1 & \textbf{91.03} & & & & & & & & & \\ \hline
2 & 89.35 & \textbf{91.68} & & & & & & & & \\ \hline
3 & 85.62 & 90.60 & \textbf{92.29} & & & & & & & \\ \hline
4 & 85.13 & 90.54 & 91.90 & \textbf{92.49} & & & & & & \\ \hline
5 & 82.65 & 88.15 & 90.85 & 92.01 & \textbf{93.06} & & & & & \\ \hline
6 & 80.41 & 87.83 & 90.41 & 91.14 & 92.77 & \textbf{92.05} & & & & \\ \hline
7 & 80.06 & 87.21 & 89.72 & 91.02 & 92.10 & 91.78 & \textbf{91.94} & & & \\ \hline
8 & 78.85 & 86.55 & 89.29 & 90.58 & 91.08 & 91.48 & 91.89 & \textbf{91.99} & & \\ \hline
9 & 78.68 & 84.51 & 88.58 & 89.87 & 90.76 & 90.91 & 91.54 & 91.74 & \textbf{91.94} & \\ \hline
10 & 78.42 & 84.00 & 88.45 & 89.38 & 90.14 & 90.69 & 91.42 & 91.59 & 91.63 & \textbf{91.56} \\ \hline
\end{tabular}
\end{table}

In the third stage, we repeated the Permuted MNIST experiment for selected $\alpha$ values with triple the number of microcolumns (from 15 to 45 microcolumns per column). We found the optimal $\alpha$ value for the 45-microcolumn configuration to be 0.01, which maintains learning plasticity across all 10 tasks (Table~\ref{tab:performance}). Table~\ref{tab:colanet_dg3} shows the degradation profile of CoLaNET with $\alpha=0.01$ and 45 microcolumns.

\begin{table}[ht]
\centering
\caption{CoLaNET degradation profile for 45 microcolumns and $\alpha=0.01$.}
\label{tab:colanet_dg3}
\begin{tabular}{|c|*{10}{c|}}
\hline
\multirow{2}{*}{Iterations} & \multicolumn{10}{c}{Tasks} \\
\cline{2-11}
 & 1 & 2 & 3 & 4 & 5 & 6 & 7 & 8 & 9 & 10 \\ \hline
1 & \textbf{93.30} & & & & & & & & & \\ \hline
2 & 91.84 & \textbf{93.23} & & & & & & & & \\ \hline
3 & 91.03 & 92.51 & \textbf{93.62} & & & & & & & \\ \hline
4 & 90.48 & 92.01 & 93.45 & \textbf{93.87} & & & & & & \\ \hline
5 & 90.19 & 91.86 & 93.02 & 93.48 & \textbf{92.97} & & & & & \\ \hline
6 & 89.75 & 91.69 & 92.88 & 93.22 & 92.86 & \textbf{93.01} & & & & \\ \hline
7 & 89.57 & 91.36 & 92.82 & 92.95 & 92.79 & 92.99 & \textbf{93.03} & & & \\ \hline
8 & 89.09 & 91.20 & 92.69 & 92.60 & 92.62 & 92.99 & 93.27 & \textbf{91.73} & & \\ \hline
9 & 89.00 & 91.19 & 92.46 & 92.57 & 92.47 & 92.92 & 92.96 & 91.69 & \textbf{92.35} & \\ \hline
10 & 88.95 & 91.29 & 92.37 & 92.37 & 92.31 & 92.65 & 92.75 & 91.43 & 92.30 & \textbf{91.22} \\ \hline
\end{tabular}
\end{table}

The CoLaNET configuration in Table~\ref{tab:colanet_dg3} learns effectively across ten tasks and resists forgetting well, degrading only $93.30 - 88.95 = 4.35\%$ on the first task after learning nine subsequent tasks (compared to 49\% degradation for the baseline model).

Permuted MNIST simplifies stability-plasticity trade-off analysis in CoLaNET by eliminating inter-task interference through pattern independence, but this artificial separation must be explicitly addressed in real-world scenarios where tasks share underlying common features.

\subsection{Two tasks based on E/MNIST}
\label{sec:experiments_emnist}

\emph{EMNIST Dataset.} The EMNIST dataset \cite{Cohen2017} extends MNIST by including both handwritten digits (0-9) and English alphabet letters (A-Z, a-z). Following \cite{Antonov2022}, we use the Balanced subset with ten selected letter classes (A, B, D, E, G, H, N, Q, R, S), resulting in 24k training and 4k test images. For two-task experiments conducted bidirectionally, we balance the tasks by using only the first 28k images from original MNIST (24k training, 4k test).

\emph{Baseline model.} We trained the identical fully connected network (512h, 10o) from Section \ref{sec:experiments_pmnist} as our baseline, achieving 95\%/93\% accuracy on MNIST/EMNIST respectively \cite{Larionov2025} on 1 epoch. The results reveal significant (75\%/69\%) forgetting in both task directions, contrasting sharply with the minimal forgetting observed in Permuted MNIST experiments (Table \ref{tab:pmnist_ann}). This substantial difference stems from the intrinsic similarity between MNIST and EMNIST images (centered character/letter patterns), which creates shared feature representations across tasks.

\emph{CoLaNET.} We conducted the E/MNIST two-task experiments with CoLaNET in two stages. Stage one tests the CoLaNET configuration from \cite{Kiselev2024b}, sweeping $\alpha$ values in Equation \ref{eq:at} from 0 to 0.03.

For the experimental pipeline, we ran ArNI-X sequentially on both tasks (forward/reverse order) plus an additional evaluation run on the first task (6 runs total). Each experiment completed within 10 minutes on a dual Quadro RTX 6000 24GB setup.

Results show that decreasing $\alpha$ increased forgetting (reaching 60\% at $\alpha=0$, similar to a basic fully connected network). Increasing $\alpha$ slightly reduced forgetting but severely hurt EMNIST performance. This happens because the tasks share features—microcolumns trained on one task also respond to the other, especially when the activation threshold is low.

\begin{table}[ht]
\small
\centering
\caption{CoLaNET accuracy results on E/MNIST in percents}
\label{tab:colanet_emnist}
\begin{tabular}{|l|c|c|c|c|c|c|}
\hline
\begin{tabular}{@{}c@{}}Parameters\\$\alpha$ and $ns$\end{tabular} & 
\begin{tabular}{@{}c@{}}MNIST\\→EMNIST\end{tabular} & 
\begin{tabular}{@{}c@{}}MNIST\end{tabular} & 
\begin{tabular}{@{}c@{}}FM1\end{tabular} & 
\begin{tabular}{@{}c@{}}EMNIST\\→MNIST\end{tabular} & 
\begin{tabular}{@{}c@{}}EMNIST\end{tabular} & 
\begin{tabular}{@{}c@{}}FM2\end{tabular} \\
\hline
0 & 88.67→82.50 & 25.09 & 63.58 & 79.72→88.65 & 19.27 & 60.45 \\
\hline
0.00125 & 92.65→82.45 & 32.32 & 60.33 & 80.57→91.37 & 30.92 & 49.65 \\
0.00125, 0 & 90.87→87.72 & 40.70 & 50.17 & 87.20→92.72 & 35.77 & 51.43 \\
0.00125, 1k & 91.87→87.50 & 43.40 & 48.47 & 87.25→92.70 & 44.80 & 42.45 \\
0.00125, 10k & 92.07→87.75 & 38.22 & 53.85 & 86.40→92.75 & 44.67 & 41.73 \\
0.00125, 100k & 91.15→87.17 & 40.39 & 50.76 & 86.62→93.45 & 44.19 & 42.43 \\
\hline
0.0015 & 92.40→82.92 & 39.90 & 52.50 & 81.15→90.10 & 35.57 & 45.58 \\
0.0015, 0 & 91.15→87.45 & 41.22 & 49.93 & 87.07→92.62 & 39.12 & 47.95 \\
0.0015, 1k & 91.97→87.90 & 41.82 & 50.15 & 88.22→93.17 & 43.92 & 44.30 \\
0.0015, 10k & 91.15→87.95 & 41.67 & 49.48 & 86.70→93.62 & 46.85 & 39.85 \\
0.0015, 100k & 92.70→88.35 & 42.90 & 49.80 & 88.10→93.52 & 48.77 & 39.33 \\
\hline
0.002 & 92.60→81.10 & 49.02 & 43.58 & 80.10→89.47 & 38.05 & 42.05 \\
0.002, 0 & 91.85→81.12 & 44.19 & 47.66 & 86.95→74.85 & 52.40 & 34.55 \\
0.002, 1k & 92.30→86.65 & 44.47 & 47.83 & 88.97→89.90 & 52.99 & 35.98 \\
0.002, 10k & 91.92→86.02 & 42.02 & 49.90 & 88.50→91.20 & 52.30 & 36.20 \\
0.002, 100k & 92.35→87.67 & 44.80 & 47.55 & 88.50→91.90 & 53.42 & 35.08 \\
\hline
0.00238 & 92.42→78.35 & 56.87 & 35.55 & 78.49→86.65 & 38.37 & 40.12 \\
0.00238, 0 & 91.10→65.52 & 52.75 & 38.35 & 83.25→29.70 & 55.95 & 27.30 \\
0.00238, 1k & 92.20→81.60 & 50.22 & 41.98 & 87.57→80.95 & 54.47 & 33.10 \\
0.00238, 10k & 92.47→79.64 & 53.02 & 39.45 & 88.07→86.70 & 58.02 & 30.05 \\
0.00238, 100k & 92.90→84.25 & 49.50 & 43.40 & 88.40→87.72 & 56.35 & 32.05 \\
\hline
0.003 & 92.12→63.95 & 66.55 & 25.57 & 74.54→82.12 & 40.24 & 34.30 \\
0.003, 0 & 91.72→26.55 & 61.00 & 30.72 & 69.17→44.40 & 44.07 & 25.10 \\
0.003, 1k & 93.35→59.77 & 63.72 & 29.63 & 82.20→36.65 & 54.72 & 27.48 \\
0.003, 10k & 91.60→44.19 & 66.87 & 24.73 & 85.92→67.62 & 56.47 & 29.45 \\
\hline
\end{tabular}
\end{table}

In the second stage of the two-task experiment, we tested the CoLaNET configuration with different numbers of virtual synapses (parameter $ns$, ranging from 0 to 100k) for various $\alpha$ values. Table \ref{tab:colanet_emnist} presents the experimental results. The FM (forgetting measure) column shows the difference between the accuracy on the first task before and after training on the second task. Specifically, the experiments demonstrate that classification accuracy on EMNIST increases significantly either with a slight decrease in $\alpha$ or with the introduction of minor renormalization (10–100k virtual synapses).

The best CoLaNET configuration maintains relatively high post-training accuracy (above 87\% on all tasks) while achieving relatively low forgetting (43\% for MNIST→EMNIST and 32\% for EMNIST→MNIST). This configuration uses $\alpha=0.023817$ and $ns=100$k. Although sacrificing some accuracy can reduce forgetting to 25\%, even this level of forgetting remains unsatisfactory in the two-task experiment.

\begin{figure}[htbp]
  \centering  \includegraphics[width=1\textwidth]{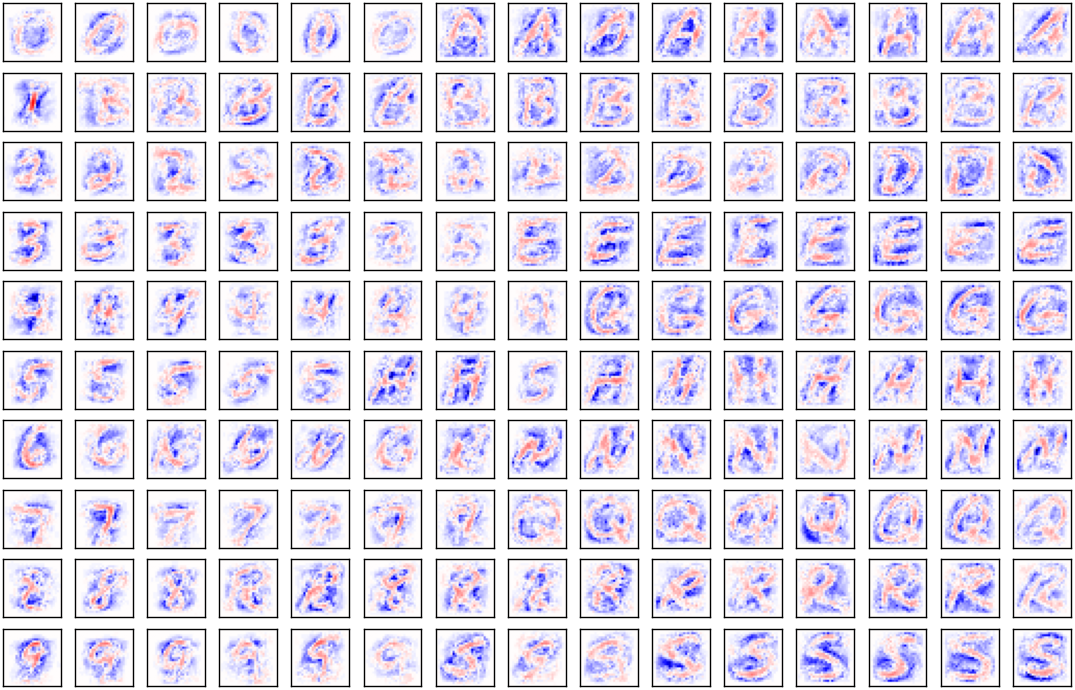}
  \caption{Receptive fields of plastic neurons after training on MNIST and subsequent training on EMNIST. Blue indicates negative weights; red indicates positive weights.}
  \label{fig:at_mnist2emnist_rf}
\end{figure}

\emph{Heatmaps.} Since CoLaNET is a fully connected network, visualizing its receptive fields becomes a convenient analysis tool. Figure \ref{fig:at_mnist2emnist_rf} displays the receptive fields (as heatmaps) of plastic neurons in CoLaNET after training on two tasks sequentially (first MNIST, then EMNIST). We observe that the columns combine patterns from both MNIST digits and EMNIST letters. For example, roughly half of the microcolumns in the first row contain a prototype of the digit 0, while the other half show a prototype of the letter A. In some columns, a small number of microcolumns experience interference between the two tasks, leading to a loss of distinct patterns—however, the vast majority of microcolumns remain specialized for a single pattern.

\subsection{Comparison of the results}
\label{sec:comparison}

Comparing CL methods is challenging due to varying datasets, problem formulations, evaluation metrics, task counts, training epochs, label assumptions, and architectures. For example, Synaptic Intelligence (SI) \cite{Zenke2017} uses 20 epochs on Permuted MNIST, while Elastic Weight Consolidation (EWC) \cite{Kirkpatrick2016} employs early stopping.

In \cite{Yin2021} replay methods are compared on Permuted MNIST using a two-layer network similar to the one discussed in Section \ref{sec:experiments_pmnist} in an online setting (one epoch). Table \ref{tab:pmnist_comparison} shows CoLaNET outperforms these methods (accounting for architectural differences).

\begin{table}[ht]
\centering
\caption{Comparison of the effectiveness of different replay approaches with a buffer size of 50 elements on ten Permuted MNIST tasks, using one training epoch per task. The bottom section presents similar results for CoLaNET in configurations with 15 and 45 microcolumns. The results are averaged over ten independent experiments.}
\begin{tabular}{|l|c|c|}
\hline
 & AA & FM \\
\hline
Joint training & $89.05 \pm 0.27$ & \\
GEM \cite{LopezPaz2017} & $74.57 \pm 0.10$ & $7.40 \pm 0.11$ \\
AGEM \cite{Chaudhry2018b} & $69.50 \pm 0.76$ & $13.10 \pm 0.63$ \\
MER \cite{Riemer2018} & $75.75 \pm 0.65$ & $8.74 \pm 0.73$ \\
MIR \cite{Aljundi2019d} & $78.31 \pm 0.63$ & $7.15 \pm 0.67$ \\
CTN \cite{Pham2021} & $79.70 \pm 0.44$ & $5.08 \pm 0.44$ \\
NCC \cite{Yin2021} & $83.47 \pm 0.43$ & $3.44 \pm 0.26$ \\
\hline
CoLaNET (15 microcolumns) & $89.67 \pm 0.16$ &  $2.75 \pm 0.23$ \\
CoLaNET (45 microcolumns) & $92.39 \pm 0.13$ &  $0.94 \pm 0.17$  \\
\hline
\end{tabular}
\label{tab:pmnist_comparison}
\end{table}

For E/MNIST, CoLaNET uses the same task set as \cite{Antonov2022} (28000 examples, 10 classes), but differs in architecture, plasticity mechanics, initialization (random values), and input encoding (DOG). Table \ref{tab:emnist_comparison} compares their results.

\begin{table}[ht]
\centering
\caption{Comparison of the effectiveness of different SNN-based approaches to transfer learning (MNIST→EMNIST) and their matching with analogous results for CoLaNET. The results in the last row are averaged over ten independent experiments.}
\begin{tabular}{|l|c|c|c|}
\hline
 & MNIST→EMNIST & MNIST & FM \\
\hline
SNN \cite{Mozafari2019} & 90.8→78.4 & 48.1 & 42.7 \\
Lateral inhibition & 94.8→88.9 & 78.6 & 16.2 \\
Pseudo-rehearsal & 93.6→79.0 & 43.5 & 50.1 \\
Self-reminder (0.25\%) & 93.6→74.0 & 74.5 & 19.1 \\
Self-reminder (10\%) & 93.6→77.8 & 91.1 & 2.5 \\
Noise regularization & 93.9→87.8 & 67.2 & 26.7 \\
Dropout & 94.2→87.6 & 62.9 & 31.3 \\
Frozen large weights & 93.6→69.2 & 78.2 & 15.4 \\
Langevin dynamics & 93.6→78.3 & 82.2 & 11.4 \\
Joint training & 93.6→79.7 & 92.0 & 1.6 \\
\hline
CoLaNET & 92.9→84.3 & 49.5 & 43.4  \\
CoLaNET + permutation & 92.14→86.96 & 90.87 $\pm$ 0.32 & 1.27 $\pm$ 0.37 \\
\hline
\end{tabular}
\label{tab:emnist_comparison}
\end{table}

The original SNN experiments in \cite{Antonov2022} show relatively high forgetting (42\%), comparable to CoLaNET (43\%). Few-shot self-reminder (memorizing 10\% of initial examples) proved their best mitigation strategy, while regularization, dropout, weight freezing, and pseudo-rehearsal proved ineffective.

In \cite{Antonov2022} is noted that interference between tasks, which leads to relatively high forgetting, occurs primarily at the deepest (third) level of the hierarchy. An experiment demonstrates this by freezing the weights of the first two layers during training on the second EMNIST task (performance on MNIST dropped by less than 1\% compared to the non-frozen version). Assuming that feature representations become less general at deeper hierarchical levels (discussed in Section \ref{sec:conclusion}), the CoLaNET architecture can effectively address this problem.

To eliminate interference between tasks, one can apply a random permutation to each task, though this sacrifices spatial patterns (discussed in section \ref{sec:experiments_pmnist}). The last row of Table \ref{tab:emnist_comparison} shows the results of CoLaNET in the configuration $\alpha=0.018$ and $ns=1000$ with an additional pixel permutation applied to each dataset. The results demonstrate that CoLaNET with random pixel permutation outperforms all other approaches, including joint training.

\section{Conclusion}
\label{sec:conclusion}

When tasks lack commonality, the optimal CoLaNET configuration remains fully resistant to forgetting in the DIL setting across many tasks. However, this configuration struggles to learn new tasks. Allowing slight forgetting (under 5\% across 10 tasks) lets CoLaNET learn new tasks effectively (with only a 1\% performance drop compared to the optimal setup).

We can control the balance between plasticity and memory stability in CoLaNET through microcolumn specialization (determined by parameter $\alpha$ in Equation \ref{eq:at}) and the number of virtual synapses (ns), which divert synaptic resource renormalization. Adding capacity by increasing microcolumns per column is possible but leads to network growth and computational overhead.

Permuted MNIST results do not fully generalize to E/MNIST due to shared features — MNIST digits resemble EMNIST letters, increasing interference as microcolumns respond to similar patterns. To mitigate this, one effective engineering approach applies random pixel permutations to each image dataset. This approach offers practical value, but has certain limitations due to disrupted spatial patterns.

Effective continual learning for tasks with shared features may require hierarchical architecture, applying CoLaNET only to deep layers. Shared features dominate shallow layers, while deeper layers exhibit task-specific representations. For example, the primary visual cortex (V1) detects low-level features (edges, lines) with high commonality, while deeper areas (V2, V4) combine them into complex, task-specific representations. Similarly, CNNs first learn general textures, then geometric shapes, and finally semantic features with minimal overlap. Hierarchy also improves capacity efficiency: shallow layers store shared knowledge, enabling FWT and avoiding redundant parameter replication.

In summary, columnar-organized SNNs provide a promising foundation for CL solutions. By combining local learning, competition, modulation, and gating mechanisms, they achieve parameter separation between tasks, balancing high plasticity and memory stability — particularly for non-overlapping feature representations, which characterize deeper layers.

\section{Conflict of interest}
This is a preprint of the Work accepted for publication in Optical Memory and Neural Networks (Information Optics), copyright 2025, https://pleiades.online.

\bibliographystyle{unsrt}  
\bibliography{main}

\end{document}